\documentclass[12pt, sort&compress]{elsarticle}
\usepackage{graphicx} 
\usepackage[a4paper, total={6.5in, 8.5in} ]{geometry}
\usepackage{amsfonts}
\usepackage{amsmath}
\usepackage{amssymb}
\usepackage{comment}
\usepackage{arydshln}
\usepackage{microtype}
\usepackage{placeins}
\usepackage[english]{babel}
\usepackage{graphicx}

\usepackage{algorithm}
\usepackage{algorithmic}
\usepackage{booktabs}

\begin{document}

\journal{Mechanical Systems and Signal Processing}

\begin{frontmatter}


\title{{Probabilistic Numeric SMC Sampling for Bayesian Nonlinear System Identification in Continuous Time}}

\author[aff1]{Joe D. Longbottom}
\author[aff1]{ Max D. Champneys}
\author[aff1]{Timothy J.\ Rogers}

    \affiliation[aff1]{
        organization={Dynamics Research Group, University of Sheffield},
        addressline={Western Bank},
        city={Sheffield},
        postcode={S10 2TN},
        country={United Kingdom}}

\begin{abstract}

In engineering, accurately modeling nonlinear dynamic systems from data contaminated by noise is both essential and complex. Established Sequential Monte Carlo (SMC) methods, used for the Bayesian identification of these systems, facilitate the quantification of uncertainty in the parameter identification process. A significant challenge in this context is the numerical integration of continuous-time ordinary differential equations (ODEs), crucial for aligning theoretical models with discretely sampled data. This integration introduces additional numerical uncertainty, a factor that is often over looked. To address this issue, the field of probabilistic numerics combines numerical methods, such as numerical integration, with probabilistic modeling to offer a more comprehensive analysis of total uncertainty. By retaining the accuracy of classical deterministic methods, these probabilistic approaches offer a deeper understanding of the uncertainty inherent in the inference process. This paper demonstrates the application of a probabilistic numerical method for solving ODEs in the joint parameter-state identification of nonlinear dynamic systems. The presented approach efficiently identifies latent states and system parameters from noisy measurements. Simultaneously incorporating probabilistic solutions to the ODE in the identification challenge. The methodology's primary advantage lies in its capability to produce posterior distributions over system parameters, thereby representing the inherent uncertainties in both the data and the identification process.

\end{abstract}

\end{frontmatter}


\section{Introduction}

Modeling in engineering serves as a crucial bridge between theoretical concepts and practical applications. It allows engineers to simulate complex systems, predict outcomes, and optimise designs, significantly reducing the need for costly and time-consuming physical prototypes. The development of these models remains a primary task, a challenge the field of system identification (SI) seeks to address \cite{ljung1998system}. 

One of the foremost challenges in SI involves fusion of theoretical knowledge with measured data to create models that not only embody theoretical understanding but also correspond with the subtleties of real-world data. When the mathematical form of a system is known and experimental data is available, a practitioner can use parametric identification as a means to merge these insights into an optimal model. 

In mechanical engineering, structural dynamic systems are often represented mathematically by second-order differential equations, based on Newton's second law of motion. These can be trivially projected into a set of first-order ordinary differential equations (ODEs) by considering the evolution of the dynamics of the state-space vector. The equations of motion for a general parameter dependent state-space model (SSM) in continuous time are familiar as,

\begin{subequations}
\begin{equation}
\dot{x}(t)=f(x(t),u(t),\theta_f)
\end{equation}
\begin{equation} \label{eq:SSMM}
y(t) = g(x(t),u(t),\theta_g)
\end{equation}
\end{subequations} 

\noindent where, $f$ and $g$ govern the transitions of the state $x$ and observation $y$ respectively, parameterised by $\theta_f$ and $\theta_g$. In the parameter identification setting, it is assumed that the functional form of $f$ and $g$ are known.

In parametric identification, the parameters $\theta_f$ and $\theta_g$ are typically determined through an optimisation procedure aimed at minimising the discrepancy between the predicted and observed system states. For linear, time-invariant SSMs, well-established techniques exist for determining unknown parameters from measured data, as detailed in \cite{viberg1995subspace, ljung1998system}. Due to the loss of superposition and challenges in analytical integration, the difficulty of parameter estimation is significantly increased for nonlinear systems \cite{worden2002nonlinearity, kerschen2006past, noel2017nonlinear}. 

Whether the system is linear or nonlinear, a critical challenge in parameter estimation arises from noise in experimental data. Such noise introduces uncertainties that complicate the optimisation process, potentially leading to biased, overconfident, or mis-specified models. If these uncertainties are not adequately addressed, they can degrade the model's precision and predictive accuracy. Bayesian parameter estimation offers a rigorous framework for probabilistic modelling, incorporating uncertainty into the estimation process, thereby enhancing model reliability in noisy environments.

Bayesian parameter estimation is well-established in structural dynamics, where it is used to quantify uncertainty by identifying probability distributions over parameters, given the noise in measurements. In a Bayesian setting, the SSM becomes probabilistic and can be written in the following Hidden Markov model: 

\begin{subequations}
\begin{equation}
      x_1 \sim p(x_1),
\end{equation}
\begin{equation}
      x_t \sim p(x_t | x_{t-1}),
\end{equation}
\begin{equation}
      y_t \sim p(y_t | x_t).
\end{equation}
\end{subequations}

\noindent where $x_t \triangleq  x(t)$ and \( p(x_1) \) denotes the prior probability distribution of the initial state, encapsulating a practitioners knowledge before any measurements are taken. The transition distribution \( p(x_t \mid x_{t-1}) \) characterises the system dynamics and inherent uncertainties, describing the evolution of the state from \( x_{t-1} \) to \( x_t \). Lastly, the measurement model \( p(y_t \mid x_t) \) represents the relationship between the current state \( x_t \) and the measurement \( y_t \), reflecting how the observed data arises from the underlying state of the system.

In the presence of uncertainty in both the system model and observations, accurate state estimation becomes a pivotal challenge to parameter estimation. Bayesian filtering and smoothing provides an optimal solution to a linear probabilistic SSM. Filtering involves the sequential update of the posterior distribution $p(x_t | y_{1:t})$ using incoming observations, while smoothing refers to the retrospective refinement of the state estimates using all available observations $p(x_t | y_{1:T})$. For an overview of such techniques see \cite{doucet2009tutorial} or \cite{särkkä2013bayesian}. For clarity the notation \( p(x_t | y_{1:t}) \) represents the posterior distribution of the state \( x_t \) at time \( t \), given all the observations \( y_{1:t} \) from time 1 to time \( t \) and $p(x_t | y_{1:T})$ represents the posterior distribution of the state \( x_t \) given all observations \( y_{1:T} \) from time \( t_1 \) to  \( t_T \).

Bayesian parameter estimation regards the model parameters as random variables. By assigning a prior distribution \( p(\theta) \), the model incorporates these parameters as conditional elements. This approach updates the probabilistic SSM to the following form:

\begin{subequations}
    \begin{equation}
        \theta \sim p(\theta),
    \end{equation}
    \begin{equation}
         x_1 \sim p(x_1 | \theta),
    \end{equation}
    \begin{equation}
        x_t \sim p(x_t | x_{t-1}, \theta),
    \end{equation}
    \begin{equation}
         y_t \sim p(y_t | x_t, \theta).
    \end{equation}
\end{subequations}

\noindent The `full' Bayesian solution requires computing the joint posterior \( p(x_{1:T}, \theta \mid y_{1:T}) \), a task which is almost always intractable. Recursive algorithms offer a viable alternative, providing a means to approximate the marginal posterior $p(\theta \mid y_{1:T})$. When parameters \( \theta \) are held constant, Bayesian recursive filtering yields the distribution:

\begin{equation} \label{eq:filter_rec}
p(\theta \mid y_{1:T}) \propto p(\theta) \prod_{t=1}^{T} p(y_t \mid y_{1:t-1}, \theta),
\end{equation}

\noindent Under this framework, a parameter estimation method was proposed by Chopin in the form of Iterated Batch Importance Sampling (IBIS) \cite{chopin2002sequential}. This approach involves the sequential sampling and reweighting of \( \theta \) values, guided by the likelihood increments \( p(y_t | y_{1:t-1}, \theta) \) as defined in  Eq.(\ref{eq:filter_rec}), with the \( \theta \)-particles being updated via resampling and Markov chain Monte Carlo (MCMC) techniques \cite{bishop2006pattern}. Chopin extends this methodology to the sequential Monte Carlo squared (SMC\(^2\)) algorithm, catering to scenarios where likelihood increments are intractable in closed form \cite{chopin2011smc2}. This method propagates multiple particle filters within the \( x \)-space, in tandem with SMC using MCMC moves in the \( \theta \)-space. These methods lay the foundation for the parameter estimation techniques employed in this research.

A critical step in all these methodologies is the evaluation of trial parameters through the unnormalised posterior likelihood, which involves comparing predicted states, contingent on the parameters, against observed data. For most nonlinear systems, the absence of a closed-form solution necessitates numerical approximation for for this step. Since the numerical solution to the system state prediction must be approximate, so must be the evaluation of the quality of the parameters made from the prediction. This means there can be no unique solution to the parameter estimation problem when numerical integration is required. Rather, `optimal' parameters identified will be dependent on the numerical error in the prediction step. Hence, the identification of parameters for nonlinear systems is intrinsically uncertain, necessitating a probabilistic approach to capture this uncertainty. This is the foundation of the probabilistic view of numerical methods, known as \emph{Probabilistic Numerics} (PN) \cite{hennig2022probabilistic}. 

PN explores the intersection of numerical analysis and probability theory, advocating for a statistical approach to numerical problems in computation. PN argues that computation, often involves solving numerical problems like linear algebra \cite{2014arXiv14022058H}, quadrature \cite{briol2017probabilistic}, optimisation \cite{hennig2013quasi} , ODEs \cite{schober2017probabilistic} and PDEs \cite{wang2021bayesian} which do not have exact solutions and thus introduce uncertainty. PN addresses this by treating numerical solvers as agents that can quantify uncertainty with probability measures, allowing for richer outputs than traditional point estimates. This probabilistic approach enables smarter, uncertainty-aware decisions within algorithms and allows for the encoding of less-than-certain expectations into solvers.

The focus of this paper is directed towards PN solutions for ODEs, which has been summarized by Tronarp et al.~\cite{tronarp2019probabilistic}. In their work, the integration of an ODE is reformulated as a Bayesian filtering and smoothing problem. Here, the proposal \(p(x_{t+1} | x_t)\) is modeled as an integrated Wiener process, and the likelihood is defined by conditioning on a \emph{pseudo-measurement} \(Z_t = z(t) \overset{\triangle}{=} 0, \, \forall t\) that captures the derivative relationship of the ODE. Consequently, the posterior \(p(x_t | y_t , Z_t)\) represents the probability of the state given the noise in the measurements and the uncertainty in the numerical integration, thus incorporating the integration uncertainty into the state estimation process.

Despite much work by many authors across the fields of uncertainty quantification in parameter estimation and numerical methods, little work has been done to combine these two areas. Schmidt et al \cite{schmidt2021probabilistic} developed a probabilistic SSM for joint inference from ODES and data able to infer the system state and latent function as a temporal Gaussian Process (GP) given the uncertainty in the data and numerical methods in a single filer update. Tronarp et al \cite{tronarp2023fenrir} developed the Fenrir algorithm for a reframing of the state and parameter estimation into a Gauss–Markov process. This methodology initiates by refining a standard Gauss-Markov prior into a physics-informed prior via PN. Subsequently, a posterior distribution is calculated using Gauss-Markov regression. Parameter estimates for the dynamical system are then extracted by maximising the marginal likelihood, resulting in a calibrated posterior distribution.

The aim of this work is to unify Bayesian parameter estimation and PN to establish a comprehensive probabilistic framework for parameter estimation in nonlinear systems. By explicitly incorporating the uncertainty in both measurement data and numerical computations, this unified approach will identify the posterior over the states given the uncertainty in the measurement and numerical integration as \(p(\theta | y_t, Z_t)\). The framework is designed to facilitate informed decision-making by providing a more complete assessment of uncertainty, thereby enabling engineers to consider the probabilistic nature of their models and to make risk-aware choices in complex engineering applications.

These aims will be achieved via the contribution of a new methodology to approximate the marginal posterior \(p(\theta | y_t, Z_t)\). The proposed methodology will be benchmarked against a set of established  nonlinear dynamic datasets. 

\subsection{Contribution}
The contribution of this work is to a nonlinear system identification method which combines the ideas of PN for solving ODEs --- as (nonlinear) filtering problems where the uncertainty in the numerical integration is quantified --- with an SMC scheme which iteratively approaches the posterior over the model parameters. %
The strength of this approach is that uncertainty from both measurement noise and the numerical methods employed is respected in the inference over the model parameters. %
Additionally, since SMC is a recursive algorithm, each intermediate posterior is a valid posterior over the parameters conditioned on an increasing subset of the time-series data; consequently, this approach provides a route to \emph{online} Bayesian estimation of nonlinear system parameters (subject to the computational requirements at each step being achievable within one sampling period). %
Finally, by explicitly including the numerical integration of the ODEs within the parameter identification procedure the methodology presented seeks to directly identify the continuous time ODE(s) which govern the nonlinear system.

\section{Probabilistic Solutions to Ordinary Differential Equations} \label{sec:PODE}

\subsection{Background}

The simulation and analysis of dynamic systems are often rooted in the concept of Initial Value Problems (IVPs), typically presented as solving ODEs with known initial conditions:

\begin{equation}
    \dot{{v}} = g(t, v(t)), \quad v(t_1) = v_1,
\end{equation}

\noindent where \( v(t)\in \mathbb{R}^d \) is the solution, \( g(\cdot)\) is a function that maps the current states to their respective derivatives, and \( v_1 \) denotes the initial states at time \( t_1 \). In a broader context, engineers are often interested in dynamic systems under forced excitation. The SSM transition function $f(\cdot)$ can be used to extend the IVP framework to accommodate forced excitation:

\begin{equation} \label{eq:SSM_trans_func}
\dot{{x}}(t) = f({x}(t),u(t),\theta), \quad x(t_1) = x_1
\end{equation}

\noindent where $x_1$ and $u(t_{1:T})$ are known. This continuous time equation allows for the exact evaluation of the derivatives of the state vector at any instance in time. Therefore, the solution to the states at any time $t$ is the integral of $f(x(t),u(t),\theta_f)$ between ${x}(t_1)$ and ${x}(t)$, 

\begin{equation}\label{eq:integral}
x(t) = \int_{t_1}^{t} f({x}(t), {u}(t), {\theta)} \, dt,
\end{equation}

\noindent When the underlying dynamic system is linear and time invariant the solution to this integral is generally available in closed form. For almost all nonlinear systems this is not the case, necessitating numerical methods. 

One such method is Euler's method, which discretises time into small increments $h$ for state progression:

\begin{equation} 
    {x}(t + h) \approx {x}(t) + h \cdot f({x}(t), {u}(t), {\theta}).
\end{equation}

Euler's method is a first order method with local error proportional to the square of $ h $ and global error linearly proportional to $ h $ such that as $h$ approaches zero the solution to Euler's method approaches the true solution to the integral \cite{davis2007methods}. However, due to computational limitations or constraints imposed by the sampled frequency of $ u(t)$ it is not always possible to reduce step size to reduce error to an negligible levels. Whilst, this can in part be combated by the use of higher order methods it can sometimes not be enough. Under these circumstances, the resulting drift in state predictions can become significant, leading to uncertainty about the true solution of the integral Eq.(\ref{eq:integral}). 

\subsection{From Ordinary to Stochastic Differential Equations}

It maybe natural therefore, to consider numerical integration as a linear SDE composed of a linear ODE to represent local linearisation,  augmented with a random variable accounting for the unknown error in the integration process. The continuous-time SDE equation is expressed as:

\begin{subequations}
    \begin{equation}
        {X}(1) \sim \mathcal{N}({\mu}(1), \Sigma(1)), 
    \end{equation}
    \begin{equation}
        d{X}(t) = [F{X}(t) + u(t)]dt + {L} d{\beta}(t),
    \end{equation}
\end{subequations}

\noindent where ${\mu}(1)$ and $\Sigma(1)$ are the mean and the covariance that describe the Gaussian distribution over the initial conditions ${X}(1)$. $F$ is the state transition matrix, $u(t)$ the force, $L$ is the diffusion matrix and $\beta(t)$ is defined as a vector of the standard Wiener processes. ${X}(t)$ is a vector of $X^{(1)}(t)$ and $q+1$ derivatives such that $X^{i+1}(t)$ is the derivative of $X^i(t)$, 

\begin{equation}
 {X}(t) =   
\begin{bmatrix}
X^{(1)}(t)\\
X^{(2)}(t)\\
\vdots\\
X^{(q+1)}(t) \end{bmatrix}
\end{equation}

\noindent It has been established that filtering and smoothing on SDEs of this nature are equivalent to GP regression, where the SDE effectively forms the GP prior \cite{sarkka2019applied}. In this context, any given realisation of the SDE corresponds to a sample from the GP prior. For foundational insights into GPs, \cite{williams2006gaussian} provides an introductory perspective. The representation of a GP as an SDE offers a notable advantage by constraining the prior to a Markov process \cite{oksendal2013stochastic}, which significantly reduces the computational complexity. Unlike general GPs that exhibit a computational complexity of \( \mathcal{O}(t^3) \), this approach reduces it to \( \mathcal{O}(t) \) \cite{hartikainen2010kalman}.

To fit the SDE in a Bayesian filtering framework measurements are required. For solving IVPs, Tronarp \cite{tronarp2019probabilistic} proposes  that a measurement can be defined upon the known derivative relationship between the states i.e. $F$, $u$ and $L$ must be set such that $X^{i+1}(t)$ is always be the derivative of $X^i(t)$. This provides the following \emph{pseudo-measurement}

\begin{equation} \label{eq:residual}
{Z} (t)=
 \begin{bmatrix}
  X^{(2)} \\
X^{(3)} \\
\vdots \\
X^{(q +1)} 
\end{bmatrix} - f \left( 
\begin{bmatrix}
X^{(1)}\\
X^{(2)}\\
\vdots\\
X^{(q)} \end{bmatrix}
\right) = 0
\end{equation}

In this framework, the SDE is conditioned on the residual relationship, as expressed in  Eq.(\ref{eq:residual}), equating to zero. For example, at any moment, the velocity must equal the derivative (as calculated from the continuous SSM Eq.(\ref{eq:SSM_trans_func})) of displacement. Any deviation from this relationship indicates an integration error. However, conditioning the process \( X(t) \) on \( {z}(t) = 0 \) for \( t \in [1, T] \) is intractable in continuous time, necessitating the adoption of a discrete time approach.

\subsection{The Discrete Time Solution}

The discrete formulation only attempts to condition the process ${X}(t)$ on $Z(t) = z(t) \overset{\triangle}{=} 0$ at a set of discrete time-points, $\{t_{1:T}\}$. Under this Bayesian filtering framework the inference problem becomes 

\begin{subequations} \label{eq:PODE}
    \begin{equation}
        {X}_1 \sim \mathcal{N}(\mu_1, \Sigma_1),
    \end{equation}
     \begin{equation}
        {X}_{t+1} | {X}_t \sim \mathcal{N}(A(h){X}_t + \xi(h), Q(h)),
    \end{equation}
     \begin{equation}
        {Z}_t | {X}_t \sim \mathcal{N}(\dot{C} {X}_t - f(C {X}_t, t_t), R),
    \end{equation}
     \begin{equation}
       {z}_t \overset{\triangle}{=} 0, \quad t = 1, \ldots, T,
    \end{equation}
\end{subequations}

\noindent where $h$ is the step size and \( R \) represents the measurement variance. In this work and in general when solving Eq.(\ref{eq:PODE}) with Gaussian filtering \( R = 0\). $C$ is the observation matrix and $\dot{C}$ is it's derivative such that \( C = 
\begin{bmatrix}
\mathbf{I} & 0 & \cdots & 0
\end{bmatrix} \)
and \( \dot{C} = 
\begin{bmatrix}
0 & \mathbf{I} & \cdots & 0
\end{bmatrix} \).
That is, \( {C} {X}_t = {X}_t^{(1)} \) and \( \dot{C} {X}_t = {X}_t^{(2)} \). It is important to observe that the likelihood model \( p(Z_t \mid X_t) \) exhibits nonlinearity, which adds complexity to the filtering solution. Additionally, \( z_t \) denotes the realisation of \( Z_t \), and the state transition \( A(h) \), process noise \( \xi(h) \), and process noise covariance \( Q(h) \) are defined as follows:

\begin{subequations} \label{eq:setup1}
\begin{align}
A(h) &= \exp(Fh), \\
\xi(h) &= \int_1^h \exp(F(h - \tau))u d\tau, \\
Q(h) &= \int_1^h \exp(F(h - \tau))LL^T \exp(F^T(h - \tau)) d\tau.
\end{align}
\end{subequations}

\subsection{Building the Model}\label{sec:build_model}

A commonly adopted prior for ODE solvers is the Integrated Wiener
Process  (IWP). Specifically, the \( q+1 \) times IWP, denoted as IWP(\( q+1 \)), is utilized due to its capability to extrapolate using polynomial splines of degree \( q+1 \) \cite{hennig2022probabilistic}. This particular choice, IWP(\( q+1 \)), facilitates the computation of integrals in Eq.(\ref{eq:setup1}) in closed form, as follows:

\begin{subequations}
\begin{equation}
A(h) = A^{(1)}(h)\otimes I,
\end{equation}
\begin{equation}
\xi(h) = 0,
\end{equation}
 \begin{equation}
  Q(h) = Q^{(1)}(h)\otimes\Gamma,
 \end{equation}
 \label{IWP_1}
\end{subequations}

\noindent where $\otimes$ is the Kronecker product, $I\in\mathbb{R}^{d\times d}$ is the identity matrix and $\Gamma\in\mathbb{R}^{d\times d}$ is a hyperparameter that calibrates the covariance in $Q(h)$.  $A^{(1)}(h)$ and $Q^{(1)}(h)$ are given by

\begin{subequations}
\begin{equation}
A(h)^{(1)}_{ij} = \mathbb{I}_{i\le j}\frac{h^{j-i}}{(j-i)!}
\end{equation}
\begin{equation}
Q(h)^{(1)}_{ij} = \frac{h^{2q+3-i-j}}{(2q+3-i-j)(q+1-i)!(q+1-j)!}
\end{equation}
\label{IWP_2}
\end{subequations}

\noindent where $\mathbb{I}$ is the indicator function.

\subsection{Filtering Solution}

Eq.(\ref{eq:PODE}) provides a Bayesian filtering problem that is linear in the prediction and nonlinear in the observation. The linear prediction step is given as, 

\begin{subequations}
\begin{align}
\mu_{t+1}^P &= A(h) \mu_t^F + \xi(h), \\
\Sigma_{t+1}^P &= A(h) \Sigma_t^F A^T(h) + Q(h),
\end{align}
\end{subequations}

\noindent where $\mu_t^F$ and $\Sigma_t^F$ are the first and second filtering moments and $\mu_t^P$ and $\Sigma_t^P$ are the first and second predictive moments.

The nonlinear observation model can be approximated using Taylor
series methods. The zeroth order extended Kalman filter (EKF0) used in this work because global uncertainty quantification can be very changing for higher order methods when the ODE is multidimensional such that d$>$1 (see \ref{sec:case_1} for an example). The filter update for the EKF0 is given as;

\begin{subequations}\label{eq:EKF0}
\begin{align}
S_t &\approx \dot{C}_t \Sigma_t^P \dot{C}^T + R, \\
K_t &\approx \Sigma_t^P \dot{C}^T S_t^{-1}, \\
\hat{z}_t &\approx \dot{C}\mu_t^P - f(C\mu_t^P, t_t), \\
\mu_t^F &\approx \mu_t^P + K_t({z}_t - \hat{z}_t), \\
\Sigma_t^F &\approx \Sigma_t^P - K_t S_t K_t^T.
\end{align}
\end{subequations}

 It is important to note that for enhanced numerical stability, the implementation of these filters using the square root formulation is recommended. A derivation of the square root Kalman filter is available in \cite{wills2012estimation}.

\subsection{Calibration}

The validity of the posterior not only depends upon the mean but also the variance $ \sigma^2$. This posterior variance is calibrated by assuring that the IWP of the GP is calibrated through the hyperparameter  $\Gamma$. Since $\Gamma$ in the EKF0 does not depend on the vector field It is proposed by Bosch et al. \cite{bosch2021calibrated} that $\Gamma$ should be a diagonal matrix $\Gamma = \text{diag}(\sigma_1^2, \ldots, \sigma_d^2)$ so that the variance can be individually calibrated for each dimension $d$. 

$\Gamma$ can be optimised by maximising the marginal likelihood, which involves selecting \( \sigma^2 \) that maximise the evidence for the observed data. The marginal likelihood is given by:

\begin{equation}
p(z_{1:T} | \sigma^2) = p(z_1 | \sigma^2) \prod_{t=2}^Y p(z_t | z_{1:t-1},\sigma^2).
\end{equation}

\noindent However, computing this is as costly as solving the ODE, thus necessitating an approximation. Bosch et al. \cite{bosch2021calibrated} proposed a quasi-ML estimator for multidimensional ODEs in the case of the EKF0, expressed as:

\begin{equation}
\hat{\Gamma}_{ii} = \frac{1}{T} \sum_{t=1}^{T}  \frac{\left(\hat{z}_{t} \right)_i}{\breve{s}_t}^2, \quad i \in \{1, \ldots, d\}.
\end{equation}

\noindent where $S_t = \Gamma\breve{s}_t $ and \( S_t \) is the innovation covariance matrix from the EKF0 equations Eq.(\ref{eq:EKF0}). This estimator, while an approximation, offers efficient calibration for the variance of the posterior and can be easily Incorporated into the EKF0 for almost no additional cost.


\section{Bayesian parameter estimation}

A fundamental aspect of Bayesian parameter identification is the ability to incorporate prior knowledge to constrain the search space of potential parameters. Using a prior enables practitioners to incorporate engineering insights into the identification process. For instance, the mass of a bridge might be approximated from its design specifications, with uncertainty accounted for by considering material property and dimensional tolerances. Even in scenarios where detailed system specifications are unknown, identification can still be bounded by known physical constraints, such as the requirement for mass, stiffness, and damping ratios to be positive. In both situations, Bayesian learning and the application of a prior allow for the incorporation of engineering knowledge into the identification process, rather than disregarding this valuable information. Consequently, Bayesian learning enhances efficiency and reduces uncertainty, while also ensuring that the identification remains confined to physically meaningful regions of the parameter space.

\subsection{Sequential Monte Carlo for parameter estimation}

Monte Carlo-based methods are numerical techniques that leverage repeated random sampling to approximate probability distributions \cite{bishop2006pattern}. Among these, the Iterated Batch Importance Sampling (IBIS) method \cite{chopin2002sequential} is designed to estimate posterior distributions of the form \(p(\theta|y_{1:T})\). The IBIS algorithm comprises two principal stages: firstly, evaluating the efficacy of a set of trial parameters \(\theta_{1:n}\), and secondly, proposing new parameters based on the assessed quality of the previous set. The innovation in IBIS is to use \emph{importance sampling} and \emph{particle rejuvenation} to explore partial distributions $p(\theta | y_{1:t})$ (where $t < T$) to efficiently explore the parameter space and evaluate the posterior $p(\theta |y_{1:T})$. This section will outline the IBIS methodology and show how it can be adapted to evaluate posteriors of the form $p(\theta | y, Z)$. The algorithm implemented in this work is shown in Algorithm 1. 

\begin{algorithm}
\caption{Particle System Algorithm}
\begin{algorithmic}[1] 
\STATE Generate a particle system $\{{\theta}_n, w_{n}\}$ that targets the prior distribution $p({\theta}_n)$
\FOR{$t = 1$ \TO $T$}
    \FOR{$n=1$ \TO $N$}
        \STATE Evaluate $p({X}_{t+1,n}\mid {Z}_t,{\theta}_n)$ as for Eq.(\ref{eq:PODE}) 
        \STATE Compute $\phi_{n}({\theta}_n) $ according to the energy function Eq.(\ref{eq:energy}) 
        \STATE Update the particle system weights in $\{\theta_{n}, w_{n}\}$ as in Eq.(\ref{eq:log_lik})
    \ENDFOR
        \IF{ESS Eq.(\ref{eq:ESS}) $<\epsilon$ }
            \STATE Sample a new particle system $\{\tilde{\theta}_{m}, \tilde{w}_{m}\}_{m=1:N}$ using an Independent Metropolis Hastings proposal, Eq.(\ref{eq:IMH})
                \FOR{$\text{s} = 1$ \TO $t$}
                    \FOR{$m=1$ \TO $N$}
                        \STATE Evaluate $p({X}_{s+1,m}\mid {Z}_s,{\theta}_m)$ as for Eq.(\ref{eq:PODE}) 
                        \STATE Compute $\tilde{\phi}_{m}({\tilde{\theta}}_m) $ according to the energy function Eq.(\ref{eq:energy}) 
                        \STATE Update the particle system weights in $\{\tilde{\theta}_{m}, \tilde{w}_{m}\}$ as in Eq.(\ref{eq:log_lik})
                    \ENDFOR
                    \STATE Accept/reject $\{{\tilde{\theta}}_m\}$ with probability given in Eq.(\ref{eq:acc_rej})
            \ENDFOR
        \ENDIF
\ENDFOR
\end{algorithmic}
\end{algorithm}

The algorithm initiates at \(t_1\), where the parameter distribution is represented by the prior \(p(\theta)\), as informed by the practitioner's engineering knowledge. Direct conditioning of \(p(\theta)\) on \(y_{1:T}\) is computationally intractable. Therefore, $p(\theta |y_{1:T})$ will be approximated by conditioning on a discrete set of random samples $\theta_{1:N}\sim p(\theta)$ (Algorithm 1.1). Using these sampled particles, initial conditions $X_{t_1,1:N}$, and the probabilistic numerical integrator (referenced in Section \ref{sec:PODE}), the distribution over the system states at subsequent time steps $({X}_{t+1}\mid {Z}_t,{\theta}_n)_{n=1:N} $ can be determined (Algorithm 1.3). 

The quality of the trial parameters can be assessed by sequentially comparing states at ${X}_{t+1}$ predicted by $f({x}(t), {u}(t), {\theta}_n)$ to the noisy measurements $y_{t+1}$ through the state space measurement model Eq.(\ref{eq:SSMM}). The comparison is made by computing the approximate energy function \cite{särkkä2013bayesian} (Algorithm 1.4), 

\begin{equation}
\phi_{t+1}(\theta) = \frac{1}{2} \log\vert 2\pi S_{t+1}(\theta)\vert +\frac{1}{2} v_{t+1}^T(\theta)S_{t+1}^{-1}(\theta) v_{t+1}(\theta)
\label{eq:energy}
\end{equation}

\noindent where $\phi$ denotes the negative log incremental likelihood or negative log incremental weight  $p(y_{t+1} | y_t, \theta_n, Z)$. The terms $S_{t+1}(\theta)$ and $v_{t+1}(\theta)$ can be calculated using the intermediate steps of the Kalman filter \cite{särkkä2013bayesian}. The energy function encapsulates the quality of the proposed parameters, considering the integration uncertainties and the influence of noisy observations. The unnormalised marginal likelihood $p(y_{t+1} | \theta_n, Z)$ can be approximated from

\begin{equation} \label{eq:log_lik}
\text{log} \, w_{t+1}(\theta_n) = \text{log}\, w_{t}(\theta_n) - \phi_{t+1}(\theta_n)
\end{equation}

\noindent where $w$ is the approximate the unnormalised weight and $w_{t=1}=1/N$ such that the update transitions the  unnormalised weight from $p(y_{t} | \theta_n, Z_{t})$ to $p(y_{t+1} | \theta_n, Z_{t+1})$ (Algorithm 1.5) and $w_{T}(\theta_n) \propto p(y_{1:T} | \theta_n, Z_{1:T}) $.  

The particle system represents \(p(\theta | y_{1:t}, Z_{1:t})\) through a weighted set of particles, rather than directly yielding a set of particles whose distribution intrinsically approximates \(p(\theta | y_{1:t}, Z_{1:t})\) itself. As the algorithm progresses through its prediction and reweighting steps (as outlined in Algorithms 1.3 to 1.5), it accumulates more information about the quality of the proposed particles $\theta_{1:N}$. This process causes the posterior distribution \(p(\theta | y_{1:t}, Z_{1:t})\) to diverge from the initial distribution \(p(\theta | y_{t=0})\). A common side effect of this process is that the particles become degenerate. This degeneracy primarily arises because, with each sequential observation, the likelihood for most particles diminishes significantly compared to a few that align closely with the observed data, resulting in an imbalance where a few particles end up with the majority of the weight. To counteract this issue, resampling and rejuvenation steps can be implemented. Resampling effectively duplicates particles with higher weights and eliminating those with lower weights so that the set of particles approximate the distribution \(p(\theta | y_{1:t}, Z_{1:t})\) its self. Therefore after resampling each particle weight $w_{1:N}$ is set to $1/N$. 

Rejuvenation, in contrast, involves the introduction of a move step to the particles post-resampling to avert the loss of diversity and facilitate the exploration of the parameter space. This process entails slightly altering each particle based on a move kernel. This move is then accepted or rejected according to a criterion that ensures the overall particle set continues to approximate \(p(\theta | y_{1:t}, Z_{1:t})\) accurately.

Efficiency in rejuvenation is critical as it is often computationally demanding, requiring a complete browsing of all past observations (Algorithm 1.9 to 1.11). Efficiency here comes from two parts. First, only rejuvenating the particles $\theta_{1:N}$ when \(p(\theta | y_{1:t}, Z_{1:t})\) is not well represented by the particles. To ensure this a standard degeneracy criterion in used,

\begin{equation} \label{eq:ESS}
ESS = \frac{(\sum_{j=1}^{N} w_j)^2 }{\sum_{j=1}^{N} w_j^2}
\end{equation}

\noindent where ESS is the effective sample size. The ESS can be compared to a user determined threshold $\epsilon$ (Algorithm 1.6). If the ESS drops below the threshold the particles are determined to be functionally degenerate and will be resampled and rejuvenated . Therefore, the threshold needs to be set at some compromise between propagating particles that can continue to infer useful information about the proposed parameters and not resampling too often and incurring unnecessary computational costs. 

The second factor to consider when managing the computational cost of the algorithm is the efficiency of the move kernel itself. Since a theoretical guarantee of convergence does not guarantee that particles will explore the search space efficiently it is key to select a move kernel with this in mind.  

For the move kernel to have a good efficiency it is important that the acceptance rate of new parameters is high. This could be easily achieved artificially via a random walk where a small perturbation to current particles is applied. However, this fails to actually rejuvenate the particles as although particles are distinct they are also very similar and therefore very highly correlated. This approach leaves degeneracy high but less detectable. 

Therefore, a move kernel should be chosen that only proposes new particles that weakly depend on the previous values. For this the independent Metropolis Hasting (IMH) kernel is chosen \cite{chopin2011smc2}. This ensures that the acceptance rate becomes better indicator of rejuvenation. 

Since the \(p(\theta | y_{1:t}, Z_{1:t})\) is approximated by the weighted particles, resampling and rejuvenation can be performed by sampling from that distribution. A rough approximation of the location of the mass of $p(\theta \mid y_{1:t}, Z_{1:t})$ is given by the expectation $\hat{E}$ and variance $\hat{V}$ of the particle system, 

\begin{equation} \label{eq:IMH}
\hat{E} = \frac{\sum_{j=1}^{N} w_j\theta_j }{\sum_{j=1}^{N_{\theta}} w_j}, \quad\hat{V} = \frac{\sum_{j=1}^{N} w_j\lbrace\theta_j-\hat{E}\rbrace \lbrace\theta_j-\hat{E}\rbrace'}{\sum_{j=1}^{N_{\theta}} w_j},
\end{equation}

\noindent such that samples can be drawn (Algorithm 1.7),

\begin{equation}
\tilde{\theta}\sim \mathcal{N}(\hat{E},\hat{V})
\end{equation}

\noindent After new parameters are proposed a complete browsing of observations from the initial time step up to the current must be performed so that the performance of the newly proposed parameters can be compared to that of the old parameters (Algorithm 1.9 to 1.11). This involves calculating the distribution over the parameters of the system given the uncertainty in the measured data and the numerical integration iteratively from time $t_1$ to $t_{t+1}$. 

With the weights of the proposed particles calculated the move is accepted with probability (Algorithm 1.12), 

\begin{equation} \label{eq:acc_rej}
\alpha = min\left( 1,\frac{p(\tilde{\theta})p(y_t\mid\tilde{\theta},Z_t)}{p(\theta) p(y_t\mid\theta,Z_t) } \right)
\end{equation}

The employment of the IMH kernel within SMC methods serves a dual purpose: resampling and rejuvenation of the particle set. This dual functionality stems from the kernel's capability to sample from the posterior distribution \(p(\theta | y_{1:t} Z_{1:t})\) as approximated by the weighted particles. Such sampling not only ensures that the distribution of particles closely approximates the posterior distribution \(p(\theta | y_{1:t} Z_{1:t})\) itself but also facilitates an effective exploration of the parameter space by the particles. Finally, the newly resampled and rejuvenated particle system is input back into the outer loop and the particle weights are reset to $1/N$ so that the process of evaluation given new observations can continue until rejuvenation is once again required. This process is repeated for each of the the discrete time steps in the data set (Algorithm 1.12).

\section{Case studies}

In this section, three case studies are presented to evaluate the effectiveness of the proposed methodology. The first case study involves a simulation of the Bouc-Wen model of hysteresis. Here, the parameter estimation by the proposed method is compared against a known ground truth, facilitating a clear assessment of its accuracy.

The latter two case studies are based on experimental datasets: the Silver Box and an Electromechanical System (EMS). These present a greater challenge in quantifying identification accuracy, due to the absence of a predetermined ground truth. However, these cases can be considered more representative of real-world system identification challenges. Their inclusion in this study is intended to demonstrate the practical applicability and adaptability of the proposed methodology in a experimental settings.

\subsection{Bouc Wen} \label{sec:case_1}

Hysteresis can be considered as a nonlinear memory effect and can be used to model internal friction, deformation of rubber and shape memory alloys to name a few examples. The equation of motion for the Bouc Wen model of hysteresis \cite{wen1976method} is given as,

\begin{subequations}
\label{eq:bouc_wen_eq_motion}
  \begin{align}
    m\ddot{x} + c\dot{x} + ky + z(x,\dot{x}) &= u(t)\\
    \dot{z}(x,\dot{x}) &= \alpha\dot{x} - \beta\left(\gamma\vert\dot{x}\vert\vert z\vert^{\nu-1}z + \delta\dot{x}\vert z \vert^\nu\right)
  \end{align}  
\end{subequations}

\noindent where, $x$ is displacement, $\dot{x}$ is velocity, $m$ is mass, $c$ is viscous damping, $k$ is linear stiffness and $z$ encodes the nonlinear hysteresis memory effect. The rate of change of $z$, $\dot{z}$ is defined by $\alpha$, $\beta$, $\gamma$, $\delta$ and $\nu$ which are used to tune the shape and smoothness of the hysteresis loop. 

For training data generation, a system based on  Eq.(\ref{eq:bouc_wen_eq_motion}) is simulated using a fourth-order Runge-Kutta scheme. The parameters for this simulation are detailed in Table \ref{tab:Bouc_Wen}. A random phase multisine load, with frequencies ranging from 0.5 Hz to 100 Hz across 2000 uniformly spaced steps, is applied to the system. The load's amplitude is initially increased linearly over the first 10\% of the simulation time, and subsequently has maximum amplitude of 208N. The simulation is conducted for 3 seconds at a sampling frequency of 131072 Hz ensuring minimal integration error. However, the only data that will be used for the identification will be acceleration measurements down sampled to 4096 Hz to provide 12288 data points. The data is then corrupted with measurement noise through the addition of \emph{i.i.d.} samples from a Gaussian distribution with zero mean and a standard deviation of 5\% of the root mean square of the acceleration.  

\begin{table} \label{tab:Bouc_Wen}
\centering
\caption{Prior parameter distributions for the Bouc Wen system.}
\begin{tabular}{ll} 
\toprule
\textbf{Prior} & \textbf{Distribution}                                    \\
               &                                                 \\
$p(m)$         & $\mathcal{N}(2.1, 0.011)$                       \\
$p(c)$         & $\mathcal{N}(8.8, 6.97)$                        \\
$p(k)$         & $\mathcal{N}(5.9\times 10^4, 2.18\times 10^8)$  \\
$(\alpha)$     & $\mathcal{N}(4.4\times 10^4, 1.74\times 10^8)$  \\
$p(\beta)$     & $\mathcal{N}(8.6\times 10^2, 6.66\times 10^4)$  \\
$p(\gamma)$    & $\mathcal{N}(0.93, 0.0541)$                     \\
$p(\delta)$    & $\mathcal{N}(1.3, 0.1056)$                      \\
\bottomrule
\end{tabular}
\end{table}

\begin{figure}
  \centering
  \includegraphics[width=0.9\textwidth]{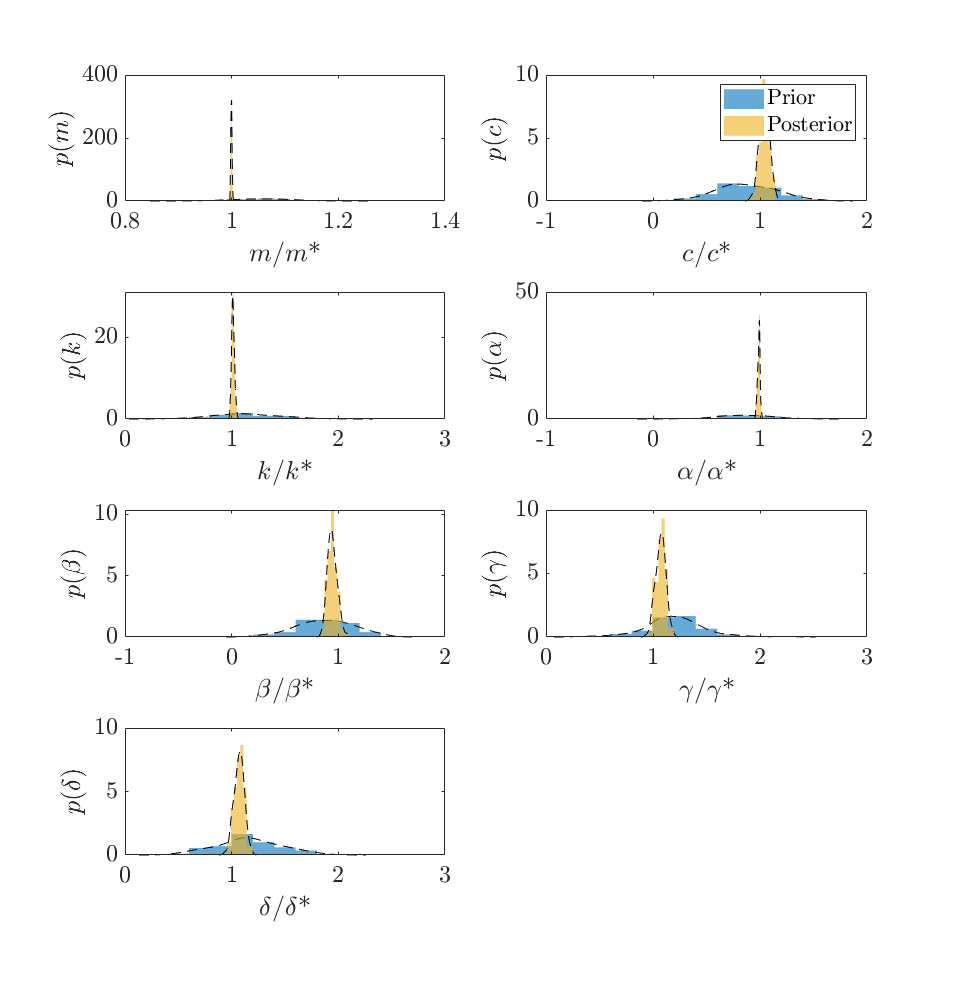}
  \caption{Prior and posterior histograms and PDFs normalised by the true parameter  values for the Bouc Wen system (denoted by $(\cdot)^*$) .}
  \label{fig:norm_prior_post_bouc_wen}
\end{figure}

Using the framework for probabilistic parameter identification outlined in this work the posterior parameter distribution is identified from a prior distribution over the parameters, the nonlinear SSM, a known forcing and a noisy measurement of the acceleration state. Figure \ref{fig:norm_prior_post_bouc_wen} shows the prior and posterior distributions over the parameters before and after training. It should be noted that Figure \ref{fig:norm_prior_post_bouc_wen} shows the parameters normalised by the ground truth so that the quality of the identification can be more easily interpreted. 

The prior means and variances can be found in Table \ref{fig:norm_prior_post_bouc_wen}. The prior was defined by perturbing the true values. Large variances are given to each of the parameters to imitated a scenario where a practitioner has low confidence in the prior values defined.

Figure \ref{fig:norm_prior_post_bouc_wen} illustrates a discernible variation in both the accuracy and uncertainty of the posterior parameter estimates. Specifically, the parameters $m$, $k$, and $\alpha$ exhibit more precise estimations with lower variance compared to the parameters $c$, $\beta$, $\gamma$, and $\delta$. This disparity is attributed to the differing extents to which each parameter influences the system's state. Consequently, it can be inferred that, under the loading conditions presented in the training data, the parameters $m$, $k$, and $\alpha$ play a more substantial role in governing the system's dynamics than $c$, $\beta$, $\gamma$, and $\delta$. A more detailed discussion on this observation is provided later in this section.

\begin{figure}
  \centering
  \includegraphics[width=1\textwidth]{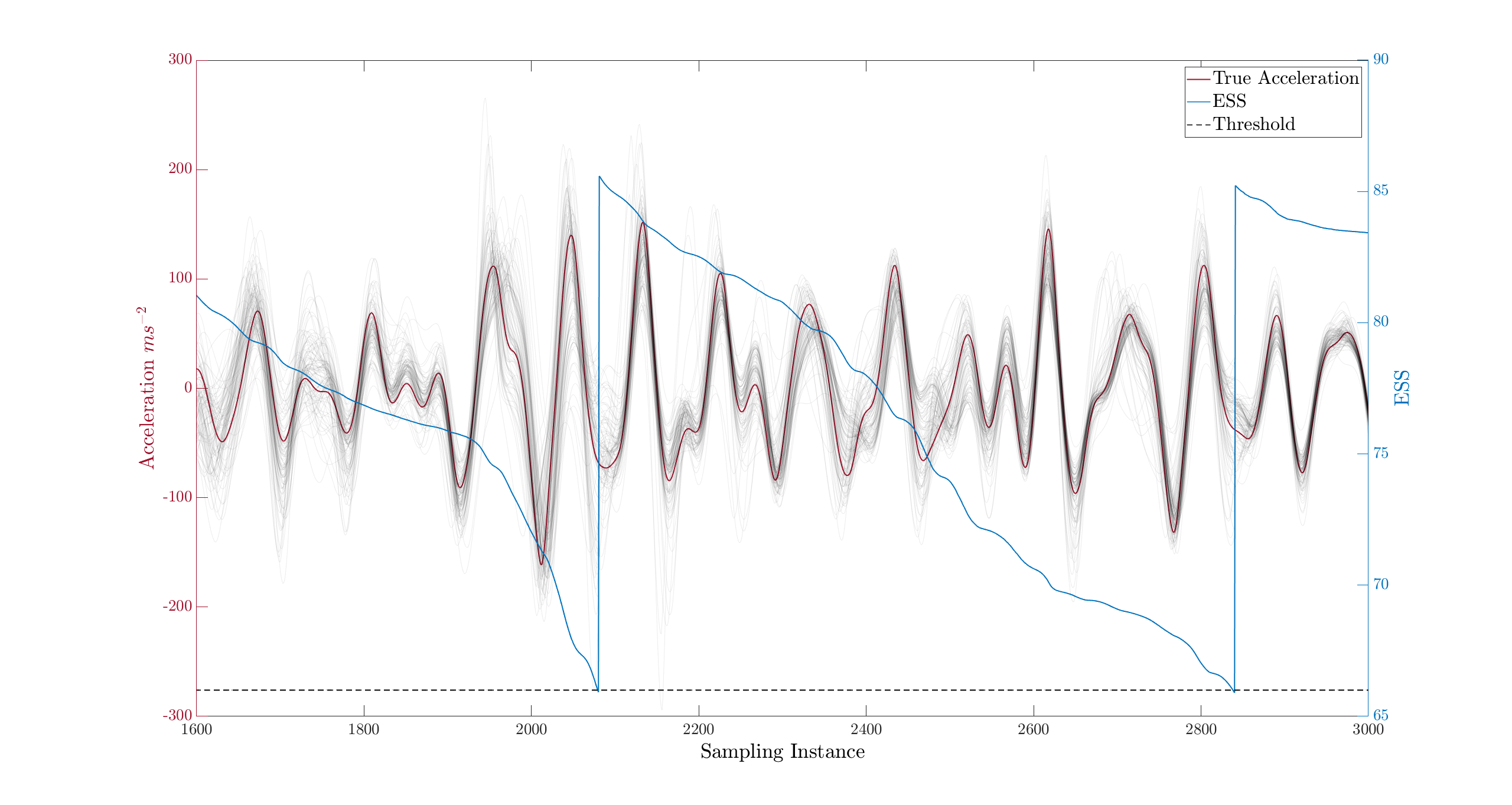}
 \caption{Acceleration samples from the filtering distribution, EES and threshold for time instances 1600 to 3000.}
  \label{fig:ESS_bouc_wen}
\end{figure}  

Figure \ref{fig:ESS_bouc_wen} serves to illustrate the phenomenon of particle degeneracy and the subsequent rejuvenation process. In this figure, the black plots shows samples from the distribution over the  acceleration given the integration uncertainty and the parameters. The effective sample size (ESS), depicted in blue, is crucial for monitoring particle degeneracy. A decline in ESS below a predefined threshold indicates functional degeneracy of parameters, necessitating their resampling.

It is important to note that the figure selectively displays only a fraction of the total resampling instances, prioritizing readability. However, the omitted data conforms to the same pattern as that which is shown. As the sampling instances progress, errors in the proposed parameters become increasingly evident, leading to a reduction in ESS until the threshold is reached. Upon resampling, the new parameter set generally exhibits closer tracking of the system's true states, with a reduced variance across the sample set. This process repeats until a complete browsing of the training data has been completed and a posterior distribution over the parameters given the entire data set is reached. 

\begin{figure}
  \centering
  \includegraphics[width=1\textwidth]{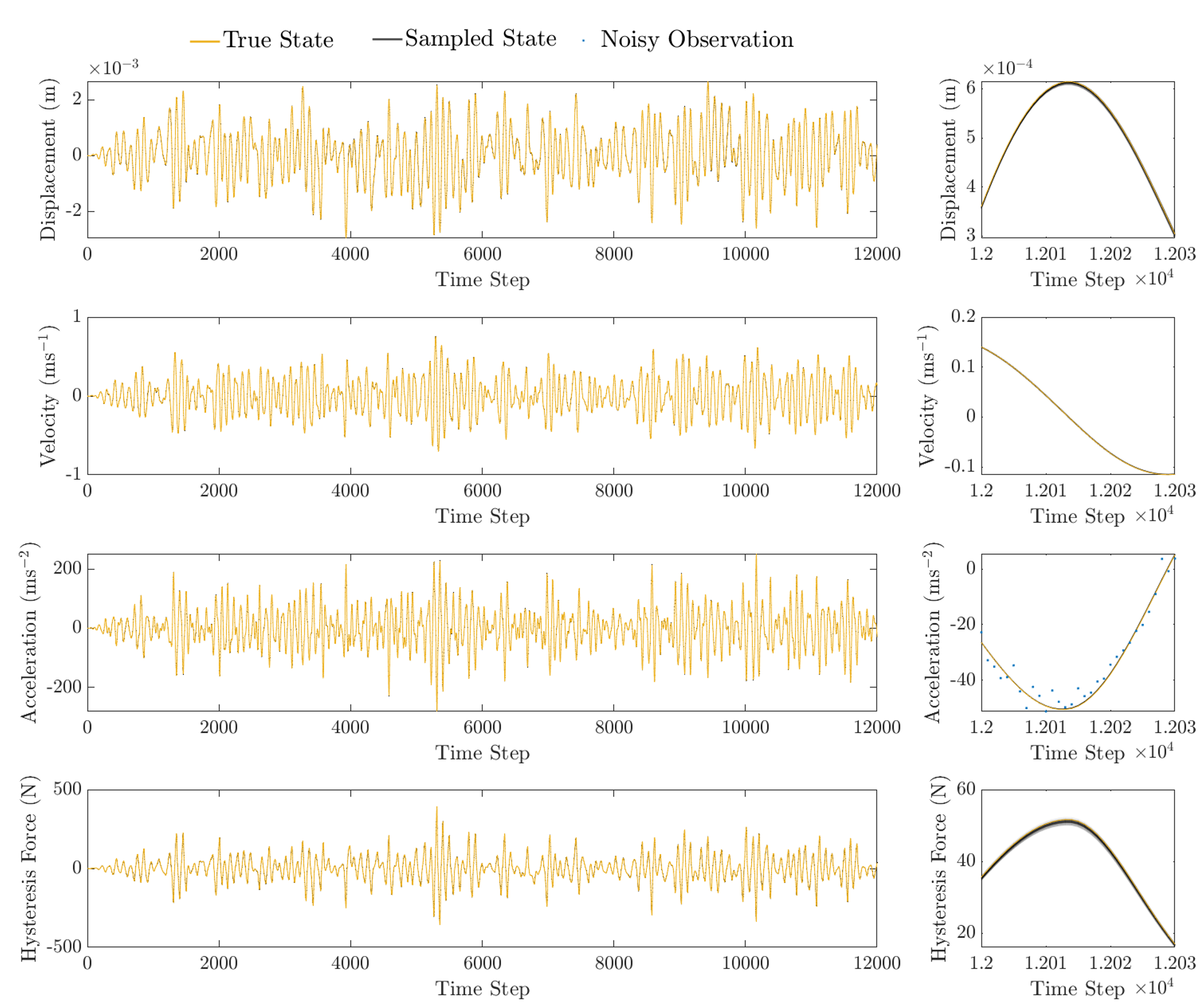}
 \caption{Samples from the posterior over the states for the Bouc Wen system plotted with the true states. An enlarge view of each state is shown to the left. Observations are plotted only for acceleration, as observations for the other states were not made available during the analysis.}
  \label{fig:true_states_bouc_wen}
\end{figure}  

Figure \ref{fig:true_states_bouc_wen} presents the true states of the Bouc-Wen system alongside the states generated from the sampled parameters of the posterior distribution. In the zoomed-in sections, it is evident that the states resulting from the posterior parameter distribution closely follow the ground truth. Additionally, it is observed that the variance of the posterior in the acceleration state is markedly lower than that of the noisy observations.  This finding indicates that the posterior distribution over the parameters accurately captures the dynamics of the Bouc-Wen system under the specified training load.

Parameter estimation is achieved through the assessment of parameter quality as an ability to predict the state of the system. Therefore, parameters can only be optimised so long as a change in a parameter has an influence in the accuracy of the predicted state. This is limited by the ability of the training data to encapsulate the dynamics of the modelled system. This is limited by the level of noise in the data, any fundamental uncertainly in the identification process such as integration uncertainty and the degree to which the data expresses the systems nonlinearity.

To see a significant further reduction in the variance of the posterior distribution further training data must be provided. This new training data must be such that a system simulated with parameters sampled from the current posterior would have a signification increase in variance of the posterior across the states. Simply put, training data must be provided that forces the states of the system outside of the current noise floor. Two different approaches may be used to achieve this. The first it to increase the number of time steps. As the number of time steps increases small inaccuracies in the parameters are more likely to cause drift in the predicted states from the true states. The second is to used training data with a greater forcing amplitude as it will make the nonlinear effect on the system dynamics more dominant. 
However, both of these approaches must be balanced against increasing integration uncertainty. For a fixed time step size increasing the number of time steps allows for integration error to accumulate and increasing forcing magnitudes increases the gradients of the states and makes the makes linear approximations within the integrator less valid. 

Here in lies a particular strength of the proposed methodology. When the uncertainly in the integration is accounted for in the posterior over the parameters it prevents the estimator from becoming overly confident in parameters that are biased due to numerical integration errors.  

For additional validation of the identified parameters, the identified parameters will be used in simulation and compared in performance to the ground truth for a benchmark testing data set \cite{noel2016hysteretic}. The benchmark data set contains noise free input and output measurements for both a sine-sweep and multisine loading condition. The data is simulated at 15000 Hz and down sampled to 750 Hz and consists of 8192 samples for both loading conditions and is performed using a Newmark integration method \cite{newmark1959method}. The sine-sweep data starts with zero initial conditions so is not steady state. The forcing amplitude is 40 N and the frequency bands rangers from 20 to 50 Hz with a sweep rate of 10 Hz/min. The random phase multi sine dataset is at steady state and the frequencies range from 5 - 150 Hz with an RMS input value of 50 N. Note that this testing data set was at no point using in the train process.

To evaluate the performance of the parameter estimation the RMSE was calculated for each particle for both the sine-sweep and a multisine loading conditions. For the sine-sweep the maximum RMSE was 6.6416$\times$10$^{-6}$ the minimum was  4.6313e$\times$10$^{-6}$ and the mean RMSE of all particles was 5.4017$\times$10$^{-6}$. For the mutisine the maximum RMSE was    6.2220$\times$10$^{-6}$ the minimum was   7.1967$\times$10$^{-7}$ and the mean RMSE of all particles was   2.4772$\times$10$^{-6}$.

\subsection{Silverbox}

The Silverbox \cite{pintelon2012system} is an electrical circuit designed to emulate the behavior of the Duffing oscillator, a single degree of freedom (SDOF) mass-spring-damper system characterised by a cubic spring term. The equation of motion for the Duffing oscillator is expressed as:

\begin{equation}
m\ddot{x} + c\dot{x} + kx + k_3x^3 = u(t)
\end{equation}

\noindent In this equation, $x$ represents displacement, $\dot{x}$ denotes velocity, $m$ is the mass, $c$ signifies viscous damping, $k$ is the linear stiffness, $k_3$ the cubic stiffness, and $u(t)$ the forcing function. While the Silver Box does not perfectly replicate the theoretical Duffing oscillator, it serves as a highly accurate approximation.

\begin{figure}
  \centering
  \includegraphics[width=1\textwidth]{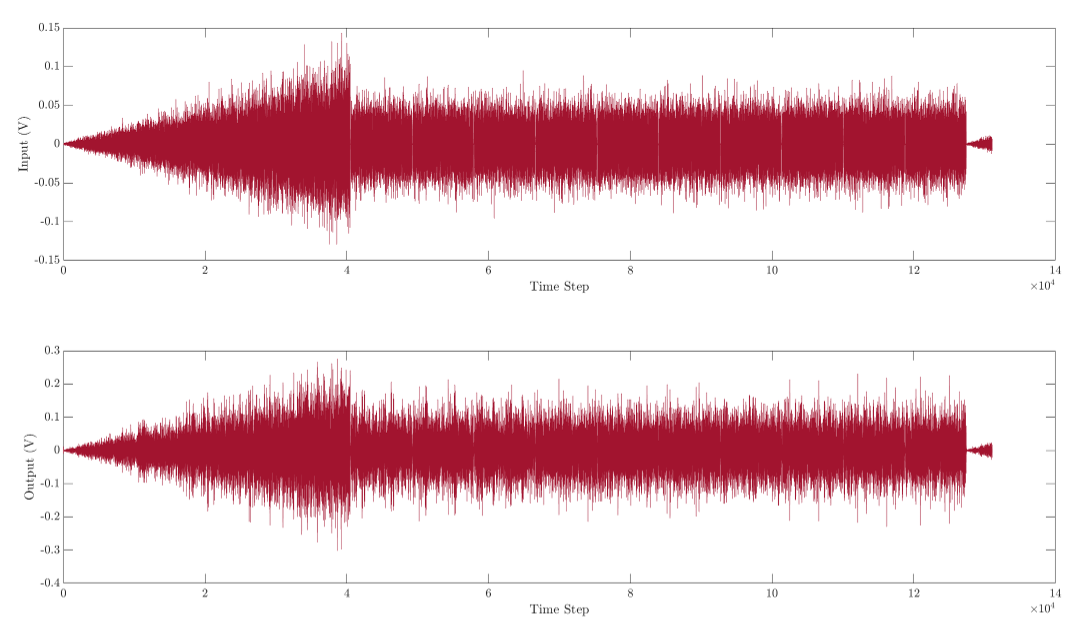}
  \caption{Measured input and response of the silver box benchmark.}
  \label{fig:silver_box_load}
\end{figure}

The Silver Box benchmark \cite{wigren2013data} encompasses two datasets. The dataset predominantly analysed, and the one chosen for analysis in this study, features an input time signal that resembles an arrow, as shown in Figure \ref{fig:silver_box_load}. In the Silver Box, both input and output are in the form of voltage. The input voltage simulates the Duffing oscillator's forcing function, while the output voltage's response to this input mimics the displacement response of the Duffing oscillator to a forcing function $u(t)$.

The input voltage shown in Figure \ref{fig:silver_box_load} is comprised of two distinct functions. The first is a Gaussian white noise of linearly increasing amplitude filtered by a 9$^{th}$ order Butterworth filter. This forms the head of the arrow and is comprised of 40,000 samples. The remaining samples shows 10 realisations of an odd random phase multisine samples at fixed amplitude. All data is recorded with a sample frequency of 610.35 Hz.

\begin{figure}
  \centering
  \includegraphics[width=1\textwidth]{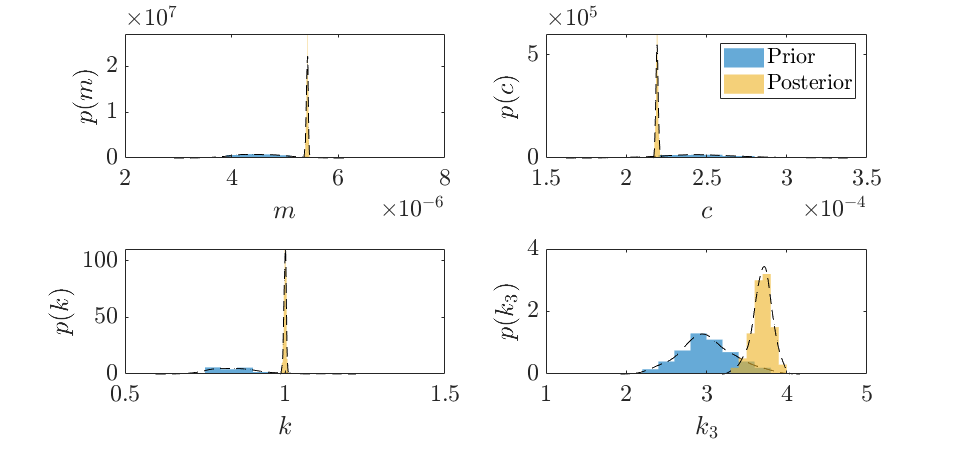}
  \caption{Prior and posterior histograms and PDFs for the silver box system.}
  \label{fig:duffing_post}
\end{figure}

The model is trained on data from the odd random phase multisine and tested on the data form the arrow head. Specifically the model is trained on 3072 observations ranging from data point 49,278 to 52,350. The prior and posterior distributions over the parameters can be found in Figure \ref{fig:duffing_post}. It can be noted that all samples of parameters are taken from a log distribution to enforce the prior that  all the parameters as defined in the Duffing equation must be positive for the identification to be physically meaningful.

Figure \ref{fig:duffing_post} illustrates the evolution of parameter distributions in the Duffing equation, transitioning from a broad prior to a more defined posterior distribution. This transition encapsulates the full extent of uncertainty inherent in the identification process. Unlike the analogous plot for the Bouc-Wen system, the parameters in this instance are not normalised against a known ground truth, as no such truth exists for this experimental system. However, an analysis of the variance ranges in the posterior distributions allows for the inference that parameters $m$, $c$, and $k$ exert a more pronounced influence on the training data compared to the cubic stiffness parameter $k_3$.
 
\begin{figure}
  \centering
  \includegraphics[width=1\textwidth]{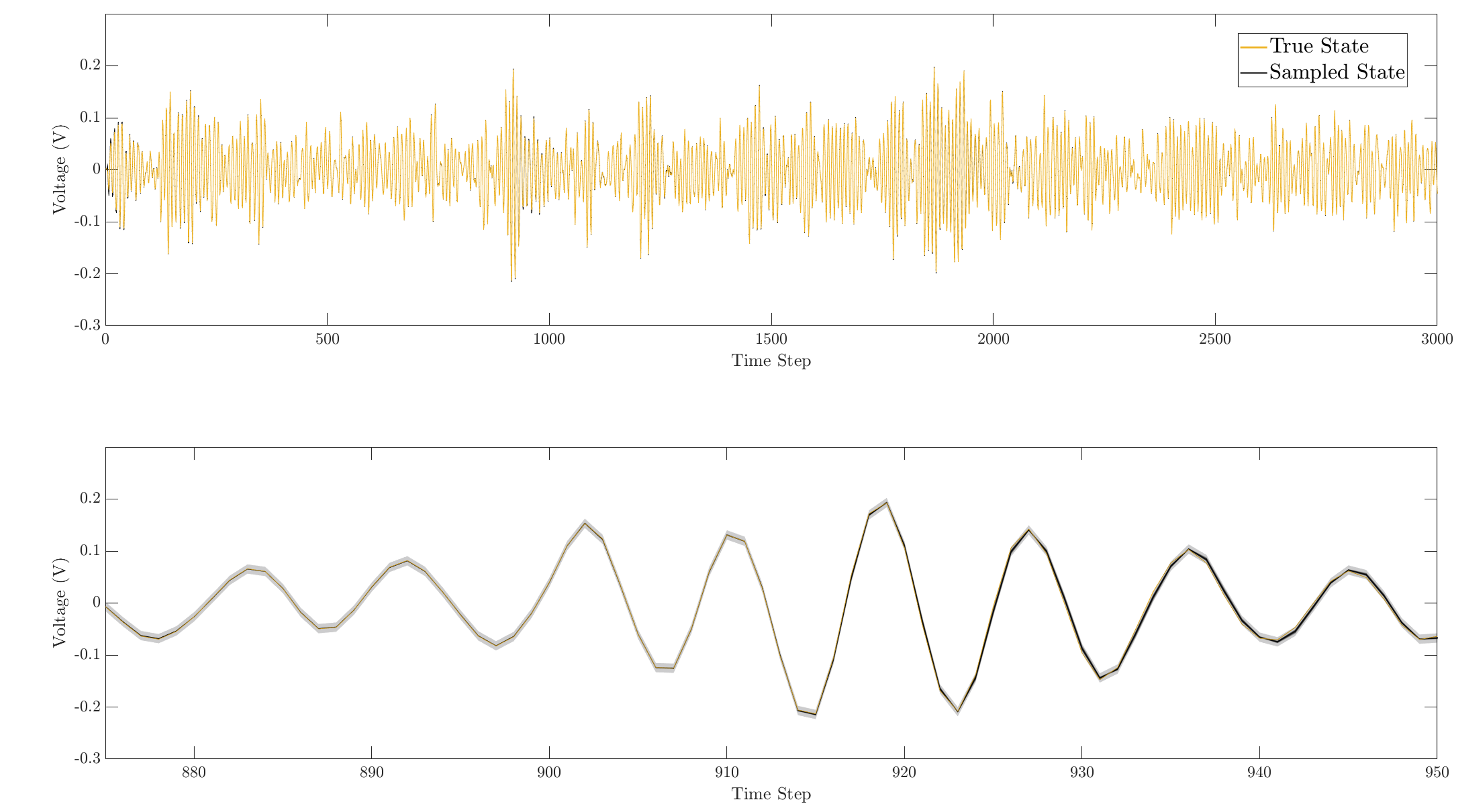}
  \caption{Samples from the posterior over the voltage for the silver box plotted with the measured output voltage. An enlarged view of the voltage is shown beneath.}
  \label{fig:duffing_states}
\end{figure}

Figure \ref{fig:duffing_states} displays the observed output voltage alongside the displacement state derived from parameters sampled from the posterior distribution. The focus is solely on the displacement state, as no observations for other states are available for comparison. This plot reveals that the sampled parameters generally align well with the observed displacement, with most observed data points residing within the sampled distribution. However, a closer examination, particularly in the zoomed-in sections, reveals instances where the observed data deviates slightly from the sampled distribution.

This deviation can be attributed to the primary sources of uncertainty in this parameter estimation. Notably, the signal-to-noise ratio for the Silver Box is sufficiently high, rendering measurement noise negligible. Consequently, during identification, the measurement noise is set to an order of magnitude around 10$^{-6}$. This leaves integration uncertainty as the dominant source of error. The observed data's lack of smoothness, as evident in Figure \ref{fig:duffing_states}, indicates that integration errors are likely to contribute significantly to the uncertainty in the states, and consequently, in the parameter quality assessment. 

Now consider how the uncertainty in the integration is calibrated. The uncertainty in the integration is calibrated using $\Gamma$ a quasi-MLE that approximates the average uncertainty from t=1 to t=T. This means at some points when the gradient of the states are small $\Gamma$ will be an over estimate and when the gradient of the states are at their largest $\Gamma$ will be an underestimate of the uncertainty in the integration. As such, it is not unexpected that at some extreme points the observed state falls outside the sampled states.  However, even considering this it should be noted that the observed states do always fall comfortably inside 2$\sigma$ of the second moment of the posterior. The second moment of the posterior is shown in faded gray in Figure \ref{fig:duffing_states}. 

Since, no ground truth is available to evaluate the quality of the identified parameters an alternative method must be used. For this the identified parameters will be used in simulation and compared to the observations for the arrowhead section of the dataset. The first 1000 time steps were excluded from this simulation to remove the transient so that the simulation was run from time step 1000 to 40,000. For this simulation a RMSE was calculated for each particle. The maximum RMSE was 2.9516$\times$10$^{-3}$ the minimum was 1.0567$\times$10$^{-3}$ and the mean RMSE of all particles was 1.8249$\times$10$^{-3}$.

\subsection{Electro-Mechanical Positioning System}

\begin{figure}
  \centering
  \includegraphics[width=1\textwidth]{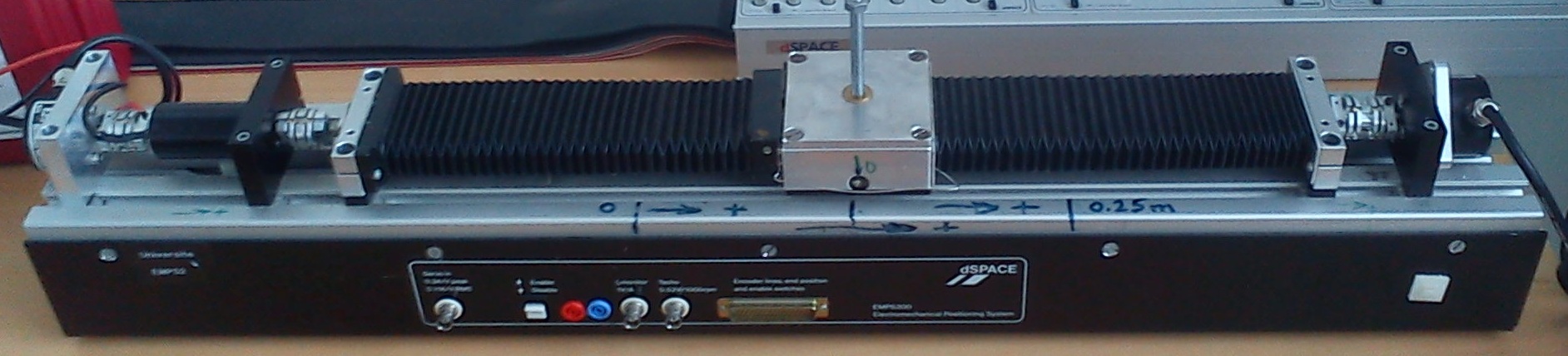}
  \caption{Image of the EMPS  \cite{janot2019data}}
  \label{fig:EMPS_image}
\end{figure} 

The EMPS (Figure \ref{fig:EMPS_image}) is a standard configuration of drive system for prismatic joints of robots or machine tools. It is comprised of a DC motor equipped with an incremental encoder and a high-precision low-friction ball screw drive positioning unit. The EMPS also features an incremental encoder and accelerometer, but their data is excluded from the benchmark, mirroring the common absence of such measurements in industrial robots \cite{khalil2002modeling}. The nonlinear equation of motion for the EMPS is given as, 

\begin{equation}
\tau_{idm}(t) = M\ddot{x}(t) + F_v\dot{x}(t) + F_csign(\dot{x}(t)) + offset
\end{equation}  

\noindent where $\tau_{idm}$ is the joint torque/force, $M$ is the inertia of the arm, $F_v$ is the viscous friction and $F_c$ is the Coulomb friction. 

\begin{figure}
  \centering
  \includegraphics[width=1\textwidth]{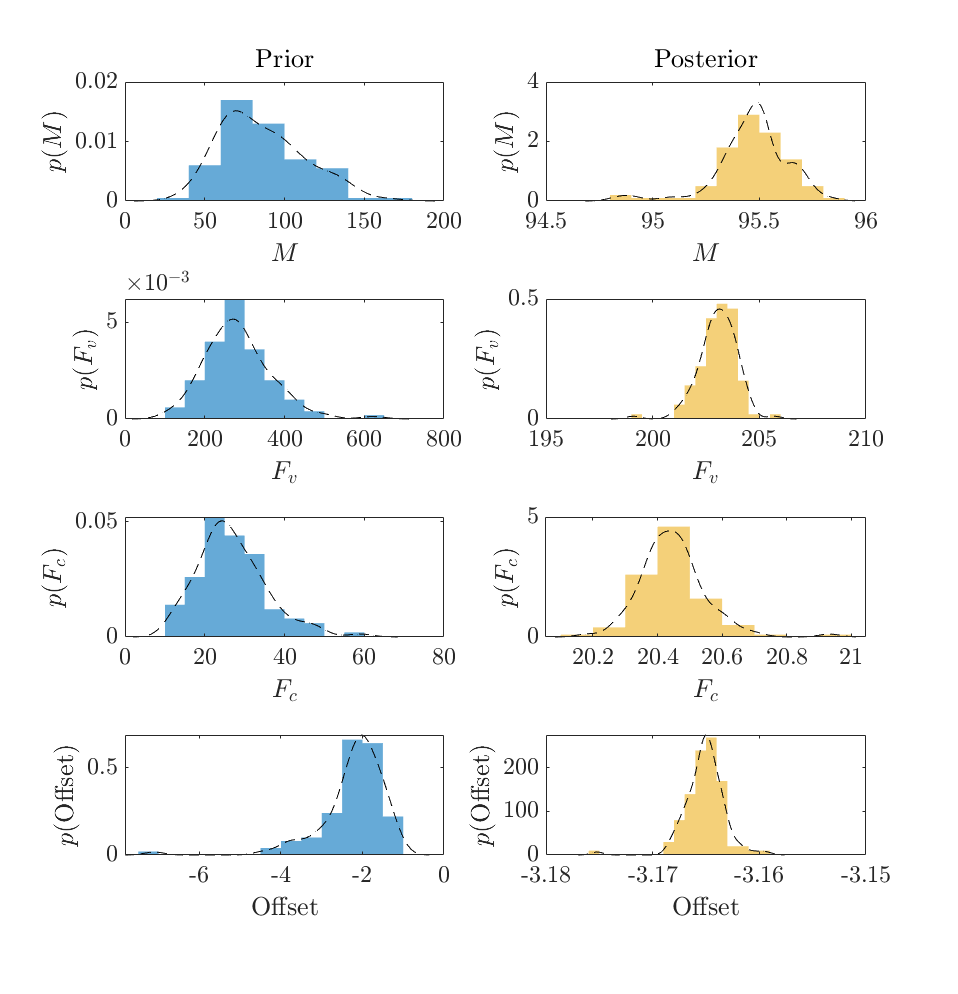}
  \caption{Prior and posterior histograms and PDFs for the EMPS shown separately due to the large reduction in variance from prior to posterior.}
  \label{fig:param_hist_EMPS}
\end{figure}

For full details of this benchmark see \cite{janot2019data}. The EMPS provides two datasets. The first is for training and the second is for testing. The training dataset consists of 25s of motor force and motor position  measurements sampled at 1000 Hz. When generating the training data the EMPS excited with bang-bang accelerations. Training is performed based on 24576 data points and the prior over the parameters is shown if Figure \ref{fig:param_hist_EMPS}. To ensure that $M$, $F_v$ and $F_c$ remain positive new proposals of these parameters will be sampled in the log space.

\begin{figure}
  \centering
  \includegraphics[width=1\textwidth]{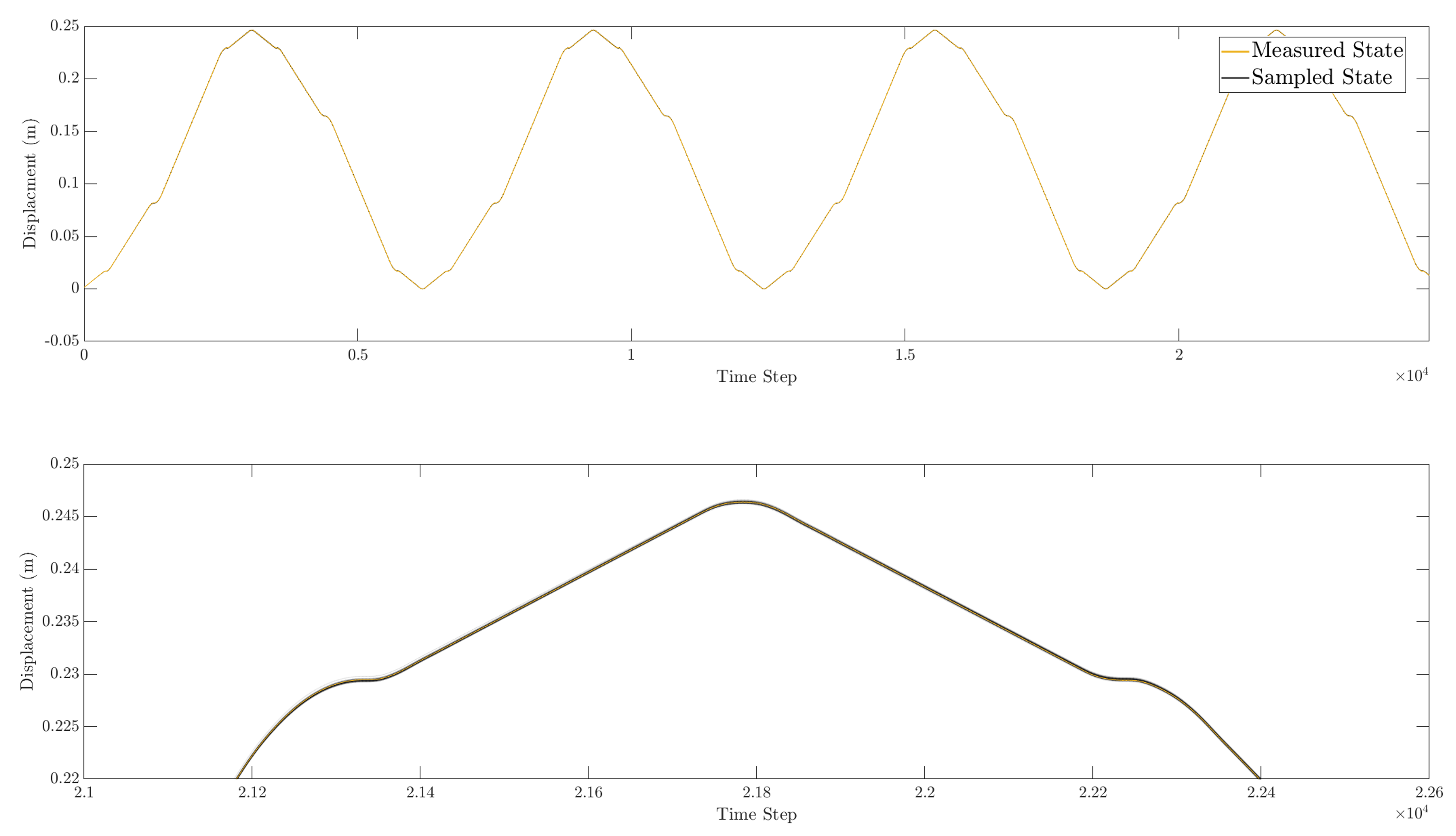}
  \caption{Samples from the posterior over the Displacement for the EMPS plotted with the measured output Displacement. An enlarge view of the displacement is shown beneath.}
  \label{fig:EMPS_states}
\end{figure}

Figure \ref{fig:param_hist_EMPS} shows the prior and posterior parameter distribution for the EMPS benchmark. It can be seen that the deffuse prior converges to a posterior with a narrow variance. Excellent convergence is seen across all four identified parameters. Figure \ref{fig:EMPS_states} shows the observed displacement plotted together with the displacement state generated from parameters sampled from the posterior distribution. Only the displacement states is shown as no observations exist for the other states. The parameters sampled from the posterior distribution correctly track the measurements with the measured states always falling within the sampled distribution.

To evaluate the quality of the identified parameters. The identified parameters will be used in simulation and compared to the observations for the testing dataset. The testing dataset consists of 25s of motor force and motor position  measurements sampled at 1000 Hz. When generating the testing data the EMPS  is again excited with bang-bang accelerations however this time there is an additional pulse component to the loading.

The identified parameters are evaluated for the for the testing dataset and compared against the measurements.  For this simulation a RMSE was calculated for each particle. The maximum RMSE was 8.9900x$10^{-4}$ the minimum was 2.9500x$10^{-4}$ and the mean RMSE of all particles was 5.2018x$10^{-4}$.

\begin{table}
\centering
\caption{Summary of Case Study Results}
\label{tab:my-table}
\begin{tabular}{@{}lllll@{}}
\toprule
\textbf{Case Study:} & \textbf{B.W. Sinesweep} & \textbf{B.W. Multisine} & \textbf{Silver Box} & \textbf{EMPs} \\
\midrule
\textbf{Unit:} & RMS (ms$^{-2}$) & RMS (ms$^{-2}$) & RMS (V) & RMS (m) \\
\midrule
Minimum Particle & 4.6313$\times$10$^{-6}$ & 7.1967$\times$10$^{-7}$ & 1.0567$\times$10$^{-3}$ & 2.9500$\times$10$^{-4}$ \\
Maximum Particle & 6.6416$\times$10$^{-6}$ & 6.2220$\times$10$^{-6}$ & 2.9516$\times$10$^{-3}$ & 8.9900$\times$10$^{-4}$ \\
Mean Particle & 5.4017$\times$10$^{-6}$ & 2.4772$\times$10$^{-6}$ & 1.8249$\times$10$^{-3}$ & 5.2018$\times$10$^{-4}$ \\
\bottomrule
\end{tabular}
\end{table}

\section{Conclusions}

This paper introduces a Bayesian parameter estimation method that unifies SMC and Probabilistic Numerics to establish a comprehensive probabilistic framework for parameter estimation in nonlinear systems.

It is the argument of the authors that: parameter estimation in nonlinear dynamic systems can abstractly be delineated into 3 sequential stages. Step 1 is to select a set of trial parameters. Step 2 involves using the set of trial parameters, the functional form of the model, and initial conditions to solve the IVP. Solving the IVP gives access to the state of the system at some future point in time. Step 3 is to compare the state pertaining to the IVP to a measured state at the same instance in time. By minimising the distance between the proposed and measured state it is possible to optimise the model parameters. 

This work adopts a comprehensive Bayesian approach throughout these 3 steps. In Step 1, a prior over the parameter space is applied. This approach reduces the size of the search space. As a result, the efficiency of the selected trial parameters is enhanced. In Step 2 a probabilistic ODE solver explicitly incorporates the uncertainty associated with solving the IVP for nonlinear dynamic systems into the identification process. Finally, Step 3 evaluates the posterior over the parameters, taking into account the uncertainties in both the integration process and the measurement.

Through 3 case studies, it was demonstrated that the proposed procedure could be realised effectively. Notable results were achieved across both simulated and experimental datasets with low RMSE. 

Nonetheless, further investigation is warranted. Future work should delve into the quantification of numerical uncertainty in more complex system identification challenges, such as an extension to multiple degree of freedom systems (MDOF) and identifying nonlinear systems where the nonlinearity's form, in addition to parameterisation, remains unknown. Moreover, extending this method to scenarios with unknown system inputs presents an open challenge deserving of further exploration.

\section*{Acknowledgements}
The authors gratefully acknowledge the support of the Engineering and Physical Sciences Research Council (EPSRC), UK through grant number EP/W002140/1 and EP/W005816/1. For the purpose of open access, the author has applied a Creative Commons Attribution (CC BY) licence to any Author Accepted Manuscript version arising. 


\bibliographystyle{unsrtnat}
\bibliography{PODE_2}
    
\end{document}